\documentclass[sigconf]{acmart}

\usepackage{graphicx}
\usepackage{subfig}
\usepackage{multirow}
\usepackage{amsmath,amsfonts}
\usepackage[linesnumbered,ruled,vlined]{algorithm2e}
\usepackage{array}
\usepackage{url}
\usepackage{bbding}
\usepackage{color}
\usepackage{colortbl}
\usepackage{xcolor}

\AtBeginDocument{%
  \providecommand\BibTeX{{%
    \normalfont B\kern-0.5em{\scshape i\kern-0.25em b}\kern-0.8em\TeX}}}

\setcopyright{acmcopyright}

\copyrightyear{2023}
\acmYear{2023}
\setcopyright{acmlicensed}\acmConference[MM '23]{Proceedings of the 31st ACM International Conference on Multimedia}{October 29-November 3, 2023}{Ottawa, ON, Canada}
\acmBooktitle{Proceedings of the 31st ACM International Conference on Multimedia (MM '23), October 29-November 3, 2023, Ottawa, ON, Canada}
\acmPrice{15.00}
\acmDOI{10.1145/3581783.3613768}
\acmISBN{979-8-4007-0108-5/23/10}

\begin{document}

\title{Universal Defensive Underpainting Patch: \\Making Your Text Invisible to Optical Character Recognition}

\author{JiaCheng Deng}
\orcid{0000-0003-0452-4031}
\affiliation{%
	\institution{Department of Computer Science, Ningbo University}
	\institution{School of Cyber Science and Engineering, Wuhan University}
	\city{Ningbo\&Wuhan}
	\country{China}
}
\email{1462492739@qq.com}

\author{Li Dong}
\authornote{Corresponding author}
\orcid{0000-0003-2002-8249}
\affiliation{%
\institution{Department of Computer Science, Ningbo University}
\city{Ningbo}
\state{Zhejiang}
\country{China}
}
\email{dongli@nbu.edu.cn}

\author{Jiahao Chen}
\orcid{0000-0002-5894-662X}
\affiliation{%
\institution{Department of Computer Science, Ningbo University}
\city{Ningbo}
\state{Zhejiang}
\country{China}
}
\email{196003641@nbu.edu.cn}

\author{Diqun Yan}
\orcid{0000-0002-5241-7276}
\affiliation{%
\institution{Department of Computer Science, Ningbo University}
\city{Ningbo}
\state{Zhejiang}
\country{China}
}
\email{yandiqun@nbu.edu.cn}

\author{Rangding Wang}
\orcid{0000-0003-2576-8705}
\affiliation{%
\institution{Department of Computer Science, Ningbo University}
\city{Ningbo}
\state{Zhejiang}
\country{China}
}
\email{wangrangding@nbu.edu.cn}

\author{Dengpan Ye}
\authornotemark[1]
\orcid{0000-0003-2510-9523}
\affiliation{%
	\institution{School of Cyber Science and Engineering, Wuhan University}
	\city{Wuhan}
	\state{Hubei}
	\country{China}
}
\email{yedp@whu.edu.cn}

\author{Lingchen Zhao}
\orcid{0000-0002-1700-3836}
\affiliation{%
	\institution{School of Cyber Science and Engineering, Wuhan University}
	\city{Wuhan}
	\state{Hubei}
	\country{China}
}
\email{lczhaocs@whu.edu.cn}

\author{Jinyu Tian}
\orcid{0000-0002-2449-5277}
\affiliation{%
	\country{School of Computer Science and Engineering, Macau University of  Science and Technology, China}
}
\email{jinyutian@ieee.org}

\renewcommand{\shortauthors}{Jiacheng Deng, et al.}

\begin{abstract}
	Optical Character Recognition (OCR) enables automatic text extraction from scanned or digitized text images, but it also makes it easy to pirate valuable or sensitive text from these images. Previous methods to prevent OCR piracy by distorting characters in text images are impractical in real-world scenarios, as pirates can capture arbitrary portions of the text images, rendering the defenses ineffective. In this work, we propose a novel and effective defense mechanism termed the Universal Defensive Underpainting Patch (UDUP) that modifies the underpainting of text images instead of the characters. UDUP is created through an iterative optimization process to craft a small, fixed-size defensive patch that can generate non-overlapping underpainting for text images of any size. Experimental results show that UDUP effectively defends against unauthorized OCR under the setting of any screenshot range or complex image background. It is agnostic to the content, size, colors, and languages of characters, and is robust to typical image operations such as scaling and compressing. In addition, the transferability of UDUP is demonstrated by evading several off-the-shelf OCRs. The code is available at \url{https://github.com/QRICKDD/UDUP}.
\end{abstract}




\maketitle

\section{Introduction}
Optical character recognition (OCR) aims to extract text from digital images, which is widely used in various commercial scenarios, such as text translation, scene text editing, and document recognition. However, powerful OCR also enables pirates to easily steal valuable or sensitive text from images, posing a threat to copyrights and privacy. 
Many websites offer non-free text services like paid novels, documents, or blogs where the text is readable but not distributable. Thus they often disable webpage JavaScript or adopt an anti-crawler mechanism to protect text copyright. 
Nevertheless, pirates can still employ OCR tools for non-intrusive attacks by capturing screenshots automatically using automated tools like Selenium and extracting the text illegally through off-the-shelf OCR software.
Once pirates disclose the paid knowledge or resell it at a low price, it will cause significant economic losses to the copyright owner. 
In order to safeguard against illegal OCR extraction while maintaining readability for humans, this work aims to develop a universal defensive underpainting patch that renders the displayed text invisible to OCR systems.

\begin{figure}[t]
	\includegraphics[width=0.8\linewidth]{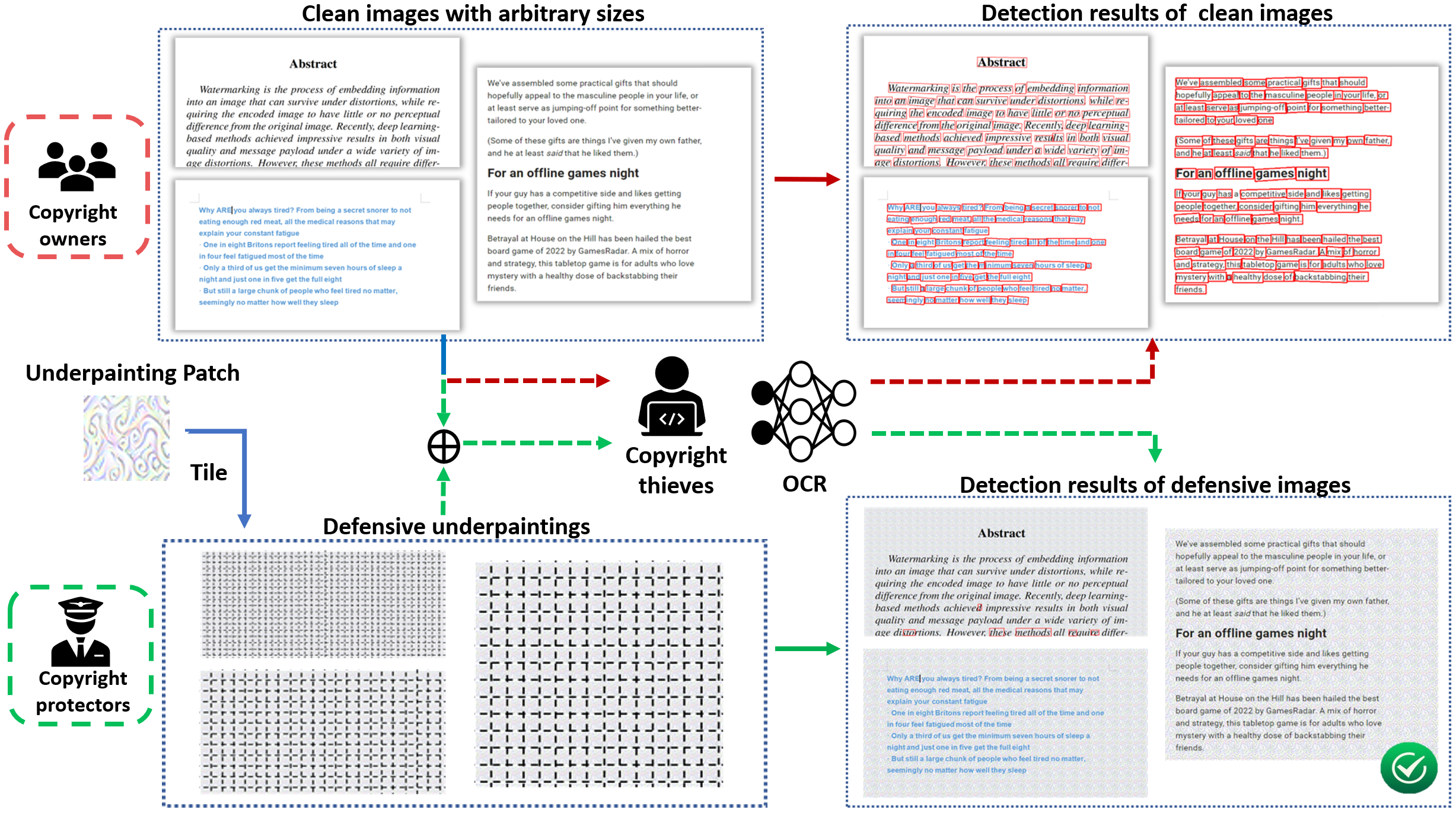}
	\caption{Illustration of proposed universal defensive underpainting patch. The underpainting patch can be tiled to arbitrary sizes and generate defensive underpainting. The proposed method can make almost all the characters undetectable and works for different colors or sizes of characters.}
	\label{fig-ocr-demo}
	\vspace{-8pt}
\end{figure}

Currently, several studies have explored the use of adversarial examples \cite{goodfellow2014explaining} to counter OCR systems. 
These works add small, imperceptible perturbations to the input image, which could cause OCR systems mis-recognize or mis-locate the text. 
Specifically, the works \cite{song2018fooling, xu2020machines} proposed targeted distortions in text areas to deceive scene text recognition models. 
Chen \emph{et al.} \cite{chen2020attacking} suggested a watermark attack that involves adding a visible watermark on the text region and distorting it to obstruct OCR. 
The works \cite{wu2021transferable,xiang2022text} proposed global and local distortions in text images to make protected text regions difficult for scene text detectors to locate.

Although these works can deceive OCR in a controlled lab environment, they encounter significant obstacles when applied to real-world scenarios. 
First, most of the existing works generate image-specific adversarial perturbations for each text image, lacking universality and being computationally inefficient for vast text images. 
Such protection schemes are impractical because one has to regenerate the adversarial perturbation when applied to different text images. 
For copyright owners, storing the text along with its corresponding perturbation incurs additional storage. 
Second, most of the existing works are unexpandable. 
Even if a universal adversarial example is successfully produced, it cannot be applied to web pages or documents with arbitrary sizes. 
Furthermore, even if protectors carefully craft protection with different sizes for every web page or document, it may be invalid due to large-scale or partial screenshots.
Third, most of the existing methods often modify character pixels to fool OCR systems and affect visual quality of texts (\emph{e.g.}, watermarking adversarial attack \cite{chen2020attacking} can fool scene text recognition but is hard for readers). 
Finally, none of these works explicitly consider robustness against image scaling operations - frequently-used operations in practical scenarios that may eliminate effects from permutation-based protections.

To tackle the aforementioned issues, we propose a Universal Defensive Underpainting Patch (UDUP) method for protecting text copyright. 
As depicted in Fig.\ref{fig-ocr-demo}, instead of distorting text characters, we suggest modifying the underpainting of a text image to make the text undetectable. 
Specifically, UDUP optimizes only a fixed-size patch, which can be applied to web pages or documents of any size through patch-tiling.
In this way, we explicitly seek a universal adversarial underpainting patch for adapting defenses to large-scale or partial screenshots in practice.
The framework of UDUP follows an iterative optimization process based on random mini-batches of text images.
Initially, we create a fixed-size patch that is tiled into the text image as the defensive underpainting. Then, this patch is optimized using a carefully designed loss function that includes prediction loss and multi-middle-layer loss components. The former guides OCR mislocation while the latter prevents overfitting to the source model. 
In addition, to ensure that UDUP protection applies to characters of different sizes and is robust to scaling, two random scaling modules are incorporated.

The main contributions of this work are summarized as follows:
\begin{itemize}
	\item For the first time, we present a defensive strategy for resisting OCR systems by only modifying the underpainting rather than the characters themselves. 
	\item We propose a universal adversarial underpainting patch for arbitrary characters, a practical text copyright protection method against OCR. The proposed adversarial patch is expandable and works well for text images of arbitrary size while resisting image scaling and JPEG compression.
	\item Extensive experiments on transferability show the effectiveness of our defense against state-of-the-art scene text detectors and publicly available commercial OCRs.
\end{itemize}

\section{Related Work}
\subsection{Scene Text Detector}
A typical OCR procedure is composed of two modules: scene text detection (STD) \cite{Baek2019CVPR,liao2020real,liao2022real} and scene text recognition (STR) \cite{sheng2019nrtr,bautista2022scene}.
STD locates the text position and return corresponding location boxes, while STR recognizes the characters in those boxes.
In this work, we focus primarily on scene text detection since accurate localization is crucial for successful character recognition. 
Traditional regression-based methods (\emph{e.g.}, EAST \cite{zhou2017east}, TextBoxes++ \cite{liao2018textboxes++}) suffer from limitations in capturing all possible shapes that exist in real-world scenarios. 
Segmentation-based approaches \cite{wang2019efficient,liao2022real,qiao2021mango,wang2021pgnet,wang2019shape} have become more widely used due to their ability to detect texts of arbitrary shape by combining pixel-level prediction with post-processing algorithms. 
As the writing of this work, the industry community is inclined to adopt the segmentation-based methods. The WeChat OCR is based on DBnet \cite{liao2020real}, and EasyOCR is based on CRAFT \cite{Baek2019CVPR}. 
We take state-of-the-art semantic segmentation models CRAFT, DBnet, PAN++\cite{wang2021pan++}, PSENet\cite{wang2019shape} and EasyOCR as exemplar attacking targets to develop UDUP, which aids in protecting copyrighted texts by helping them escape detection from STDs.

\begin{figure*}[t]
	\centerline{\includegraphics[width=0.82\linewidth]{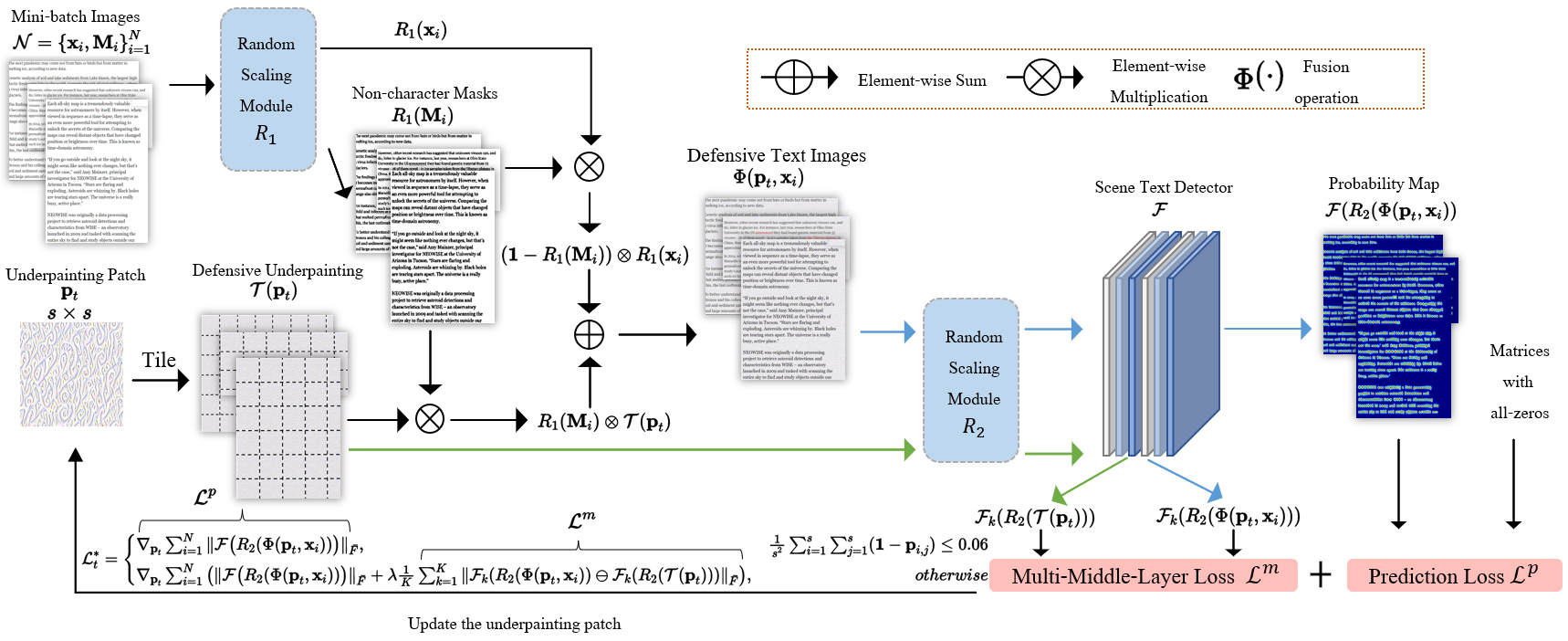}}
	\caption{
		The proposed universal defensive underpainting patch for evading scene text detector of OCR.
		Each iteration is based on random mini-batch text images.
		First, $R_1$ rich character size by randomly scaling the clean input image.
		Then we fuse the text with the defensive underpainting generated by tiling the patch.
		Next, $R_2$ improves the robustness of scaling for UDUP via randomly scaling defensive text images.
		Finally, multi-middle-layer loss $\mathcal{L}^m$ and prediction loss $\mathcal{L}^p$ be calculated, and the underpainting patch is updated.
	}
	\label{fig-ocr-framework}
\end{figure*}

\subsection{Adversarial Attacks on OCR}
Current adversarial attacks on OCR aim to deceive recognition results and make text undetectable. Previous works \cite{xu2020machines,chen2020attacking,song2018fooling} introduced subtle distortions in text areas to mislead STR models. 
Xiang \emph{et al.} \cite{xiang2022text} added distortion to the text area, rendering the text undetectable.
However, these methods are not universally applicable as they require individual distortion for each text image. 
Wu \emph{et al.} \cite{wu2021transferable} proposed universal adversarial examples to evade scene text detection but limited their usage by requiring fixed-sized input images of texts, which is impractical for real-world scenarios where web pages or documents vary in size and screenshots can be arbitrary.
Moreover, none of these approaches considered the robustness of scaling and compression.

\section{Universal Defensive Underpainting Patch}
\label{sec:UAUP}
This section presents the proposed method for generating a universal defensive underpainting patch to defend against STD. An overview of the framework is shown in Fig.\ref{fig-ocr-framework}, such a defensive patch can be tiled to a text image of arbitrary size, creating a defensive underpainting that could paralyze a text detector. In the next, we first provide mathematical problem formulations and then give a detailed explanation for each module.

\subsection{Problem Formulation}
Given $\mathbf{M}_i$ as the mask for indicating the non-character pixel region. 
Let $\mathcal{D}=\{\mathbf{x}_i,\mathbf{M}_i\}^N_{i=1}$ be a dataset of $N$ text images, where $\mathbf{x}_i\in[0,1]^{h_i\times w_i}$ is a text image of size $h_i\times w_i$. We shall emphasize that the height $h_i$ and width $w_i$ are arbitrary and not necessarily the same, considering the complex application scenario of an OCR system. 
Given an STD model $\mathcal{F}$, the output of $\mathcal{F}(\mathbf{x}_i)$ is a probability map $\mathbf{y}_i\in[0,1]^{h_i\times w_i}$, denoting the probability that each pixel belonging to text region. As shown in Fig.\ref{fig-ocr-framework}, the goal of UDUP is to seek a universal defensive patch $\mathbf{p}^*\in[1-\epsilon,1]^{s\times s}$, where $\epsilon$ is the parameter restricting the perturbation magnitude and  $s$ is the size of UDUP patch. Note that the patch size is much smaller than the size of the image, \textit{i.e.}, $s \ll \min(h_i, w_i)$.
More formally, the seek of UDUP patch $\mathbf{p}^*$ can be formulated as the following optimization problem,
\begin{equation}\label{equ:1}
	\begin{aligned}
		\mathbf{p}^*= &\underset{\mathbf{p}}{ \operatorname {arg~min~} } \mathbb{E}_{(\mathbf{x},\mathbf{M}) \sim \mathcal{D}} \left[\mathcal{L}^p(\mathbf{x},\mathbf{p})+\lambda \mathcal{L}^m(\mathbf{x},\mathbf{p})\right]\\
		&\text{s.t.:}~ \Vert \mathbf{1}-\mathbf{p} \Vert_{\infty} < \epsilon,
	\end{aligned}
\end{equation}
where $\mathcal{L}^p$ and $\mathcal{L}^m$  are prediction loss and multi-middle-layer loss, respectively. $\lambda$ is the weighting hyper-parameter, balancing the importance between $\mathcal{L}^m$ and $\mathcal{L}^p$.

Upon obtaining the UDUP patch $\mathbf{p}^*$, one can \textit{tile} such patch in a non-overlapping fashion towards the specified size $h_i\times w_i$ for the $i$-th image, forming the defensive underpainting that could mislead the STD. Mathematically, for the $i$-th image, its \textit{tiling} operation is a function $\mathcal{T}_i : \mathbb{R}^{s\times s} \rightarrow \mathbb{R}^{h \times w}$ which can be defined as follows
\begin{equation}
	\mathbf{u}_i = \mathcal{T}_{i}(\mathbf{p}^*),
\end{equation}
where $\mathbf{u}_i \in \mathbb{R}^{h_i\times w_i}$ is the defensive underpainting, whose $(m,n)$-th element can be computed by
\begin{equation}
	\mathbf{u}_i[m,n]  = \mathbf{p}^*[ m \% s, n \% s]
\end{equation}
where $m$ and $n$ represent the coordinates of the pixel. 
$0\leq m \leq h_i-1 $, $0 \leq n \leq w_i-1$, and $\%$ denotes the modulus operation.

However, solving the optimization problem (\ref{equ:1}) over the entire dataset is computationally difficult due to the huge number of samples. To alleviate this issue, inspired by \cite{shafahi2020universal}, we suggest employing a gradient descent-based method to find the UDUP $\mathbf{p}^*$ through multiple iterations, using mini-batches $\mathcal{N}$ of samples rather than the entire dataset in each iteration.
Denoting $\mathbf{p}_t$ and $\mathcal{L}^*_t$ as the UDUP and the loss at $t$-th iteration, the original optimization problem (\ref{equ:1}) can be recast as
\begin{equation}\label{equ:2}
	\mathcal{L}^*_t=\underset{(\mathbf{x},\mathbf{M}) \sim \mathcal{N}}{\mathbb{E}}[ \mathcal{L}^p(\mathbf{x},\mathbf{p}_t)+\lambda \mathcal{L}^m(\mathbf{x},\mathbf{p}_t)],
\end{equation}
\begin{equation}\label{equ:3}
	\mathbf{p}_{t+1}=\text{Clip}_\epsilon\{\mathbf{p}_t+\alpha \cdot \text{sign}(\nabla_{\mathbf{p}_t}\mathcal{L}^*_t)\},
\end{equation}
where $\text{Clip}_\epsilon\{\cdot\}$ clips the input variable such that $\Vert \textbf{1}-\mathbf{p}_{t+1} \Vert_{\infty} < \epsilon$. The $\mathbf{p}_0$ is initialized as a matrix with all elements populated with ones.

\subsection{Random Scaling Modules}\label{sec:diversity}
As shown in Fig.\ref{fig-ocr-framework}, the framework of UDUP consists of two random scaling modules, \textit{i.e.}, $R_1$ and $R_2$. 
In fact, both of these two random scaling modules operate image resizing. For notational simplicity, we use $R_1$ and $R_2$ to differ that the two random scaling modules may involve different scaling factors.

Formally, the image scaling operation can be generally expressed as $R_r(\mathbf{x})$, where $r$ is the scaling factor that controls the output image is $r$ times the size of image $\mathbf{x}$. Here, the scaling factor $r$ is randomly drawn from the uniform distribution $\mathcal{U}(a,b)$. The model parameter $a$ and $b$ can be determined by
\begin{equation}\label{equ:d}
	a=\max\left(0.9^{\lceil t/\beta \rceil},0.6 \right),~
	b=\min\left(1.1^{\lceil t/\beta \rceil},2 \right),
\end{equation}
where $t$ is the number of iteration, $\beta$ is a hyper-parameter, which is empirically set $6$ in the experiments. 

Remark that the initial ranges of all scaling operations are small and gradually expand.
This is to prevent overly strong warping from retarding the training process. Although the two random scaling modules implement scaling with the same image resizing strategy, the underlying motivations beneath are quite different.
The motivation for designing $R_1$ is to enrich the diversity of the data for training the universal defensive patch. This is based on the observation that, in real-world scenarios, the size of characters displayed on a text image varies widely and UDUP shall be character-size resilient. 
Note that a similar data-diversify strategy was already practised in several previous works \cite{xie2019improving,athalye2018synthesizing}.
In contrast, the motivation of $R_2$ is different. Scaling web pages or documents is a frequent operation of a pirate, requiring UDUP to own the anti-scaling capability. To this end, the module $R_2$ is incorporated to randomly scales defensive text images, mimicking the pirate attacks on the protected text images. 

\subsection{Loss function}\label{sec:loss}
The overall loss $\mathcal{L}^*$ consists of prediction loss $\mathcal{L}^p$ and multi-middle layer loss $\mathcal{L}^m$.
Next, we introduce the motivations for designing the two losses, respectively.
For ease of the notional expression, the element-wise addition, subtraction and multiplication operations are denoted by $\oplus$, $\ominus$ and $\otimes$, respectively.

\noindent\textbf{Prediction Loss.} UDUP aims at adding a defensive underpainting generated by patch-tilling to help text evade detection. That is, to minimize the probability that the text area to which the underpainting is added is correctly detected. 

We denote $\Phi(\mathbf{x}_i,\mathbf{p})$ as a function to fuse the defensive underpainting $\mathcal{T}(\mathbf{p})$ and text $R_1(\mathbf{x}_i)$, resulting in a defensive text image. $\Phi(\mathbf{x}_i,\mathbf{p})$, which can, be formulated as:
\begin{equation}\nonumber
	\small{
		\Phi(\mathbf{x}_i,\mathbf{p})=\big(R_1(\mathbf{M}_i)\otimes \mathcal{T}_i(\mathbf{p})\big)\oplus\big((\mathbf{1}-R_1(\mathbf{M})_i)\otimes R_1(\mathbf{x}_i)\big),
	}
\end{equation}
where $R_1(\mathbf{M}_i)\otimes \mathcal{T}_i(\mathbf{p})$ represents a defensive underpainting that does not contain the text pixels, $(\mathbf{1}-R_1(\mathbf{M})_i)\otimes R_1(\mathbf{x}_i)$ represents a text image without underpainting. Here $\mathcal{T}_i(\mathbf{p})$ tiles the patch to the defensive underpainting.

Then, the prediction loss is defined as follows:
\begin{equation}
	\mathcal{L}^p(\mathbf{x}_i,\mathbf{p})=\Vert \mathcal{F}\big(R_2(\Phi(\mathbf{x}_i,\mathbf{p}))\big)\Vert_{\bar{F}},
\end{equation}
where $\Vert \cdot \Vert_{\bar{F}}$ denotes a variant of Frobenius Norm that is defined as follows. For a matrix $\mathbf{x}$ of the size $h\times w$, $\Vert \mathbf{x} \Vert_{\bar{F}}$ can be computed by
\begin{equation}
	\small
	\Vert \mathbf{x}\Vert_{\bar{F}}=\frac{1}{hw} \sum_{i=1}^H\sum_{j=1}^W (\mathbf{x}[i,j])^2.
\end{equation}

It is worth noting that incorporating $\mathcal{T}_i(\cdot)$ into the loss is of utmost importance. Firstly, this operation enhances the ability of the patch to counteract detection for various characters and increases the universality of UDUP. 
Secondly, the tiling operation could trigger adversarial behaviour even for a local patch, making the UDUP effective for arbitrary-size screenshots that a pirate may capture.

\begin{algorithm}[t]
	\caption{Universal Defensive Underpainting Patch}
	\label{alg:algorithm}
	\KwIn{Training text image dataset $\mathcal{D}$, the target victim STD model $\mathcal{F}$. The max number of iterations $T$, allowed perturbation magnitude $\epsilon$,step size $\alpha$, patch size $s$, weighting parameter $\lambda$, the decay of momentum $\mu$.}
	\KwOut{Universal defensive underpainting patch (UDUP) $\mathbf{p}_T$ of size $s\times s$.}
	
	Initialize $\mathbf{p}_0=\textbf{\text{1}}^{s\times s}$;\\
	Initialize $\mathbf{g}_0=\textbf{\text{0}}$;\\
	\For{$t=1...T$}{
		Extract a mini-batch image set $\mathcal{N}$ from $\mathcal{D}$;\\
		Calculate MUI via (\ref{eq-mui});\\
		\eIf{\rm{MUI}$\ge0.06$}{
			$\mathcal{L}^*_t\leftarrow\sum_{\mathbf{x}_i \in \mathcal{N}}\big(\mathcal{L}^p(\mathbf{x}_i,\mathbf{p}_t)+\lambda\mathcal{L}^m(\mathbf{x}_i,\mathbf{p}_t)\big)$;\\}{$\mathcal{L}^*_t\leftarrow\sum_{\mathbf{x}_i \in \mathcal{N}}\mathcal{L}^p(\mathbf{x}_i,\mathbf{p}_t)$;\\}
		$\mathbf{g}_{t+1}=\mu\cdot \mathbf{g}_t+\frac{\nabla_{\mathbf{p}_t}\mathcal{L}^*_t}{\Vert \nabla_{\mathbf{p}_t}\mathcal{L}^*_t\Vert_1}$;\\
		$\mathbf{p}_{t+1}=\text{Clip}_\epsilon\{\mathbf{p}_t+\alpha \text{sign}(\mathbf{g}_{t+1})\}$
	}
	\textbf{return} $\mathbf{p}_T$
\end{algorithm}

\noindent\textbf{Multi-middle-Layer Loss.}
Previous works \cite{zhao2019seeing,huang2019enhancing,zhang2022improving} have found that adversarial examples based on the final predicted probability map may overfit the architecture or feature representation of the source model.
To tackle this issue, we propose to incorporate a multi-middle-layer loss into the optimization objective. 

Previous middle-layer attacks aim to maximize the difference in features between perturbed and clean images. However, this approach is not suitable for avoiding STDs, as explained in this study. The issue arises from text images having both text and non-text regions sharing the same underpainting, making it challenging to maximize the difference between defensive text images and clean images without generating conflicting optimization goals. For example, increasing the difference of features between clean and perturbed images in non-text regions may result in them being mistaken for text, while increasing the corresponding difference in text regions may render text undetectable, creating two different optimization goals. 

\begin{figure}[]
	\centerline{\includegraphics[width=1.0\linewidth]{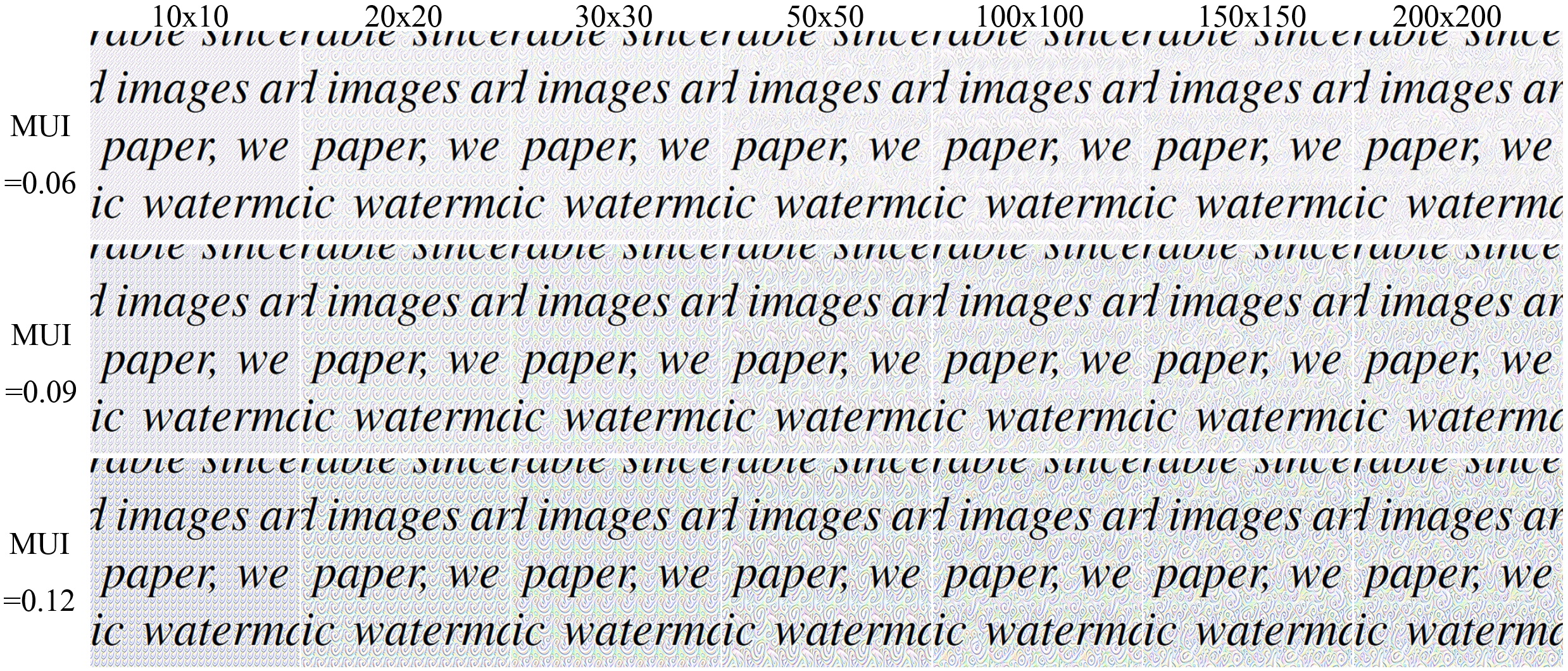}}
	\caption{Comparison of the visual quality of text images with defensive underpainting for various patch sizes and MUIs. Zoom in for better visualization. ($\lambda$ fixed as 0)}\vspace{-0.2cm}
	\label{fig-visual}
\end{figure}

Therefore, $\mathcal{L}^m$ is designed to minimize the distance between the defensive text image and the defensive underpainting, which can be expressed as
\begin{equation}\label{equ:lp}\nonumber
	\small
	\mathcal{L}^m(\mathbf{x}_i,\mathbf{p})=\frac{1}{K} \sum_{k=1}^K \Vert \mathcal{F}_k(R_2(\Phi(\mathbf{x}_i,\mathbf{p}))\ominus \mathcal{F}_k(R_2(\mathcal{T}_i(\mathbf{p})))\Vert_{\bar{F}},
\end{equation}
where $K$ is the number of selected middle layer(s) and $\mathcal{F}_k(\cdot)$ represents the output of $k$-th layer of the model $\mathcal{F}$. Note that the $\mathcal{L}^m$ loss is only used when the underpainting patch has a reliable intensity. This is because the text image contains a large number of non-text areas and the defensive performance of the underpainting patch produced by the initial iteration is very weak. At this time, minimizing the distance between the underpainting and the defensive text image will make the underpainting patch inclined to generate a blank patch, thus conflicting with the loss function $\mathcal{L}^p$. Specifically, we define the mean underpainting intensity (MUI) as:
\begin{equation}\label{eq-mui}
	\small
	\text{MUI}=\frac{1} {s^2} \sum_{i=1}^s\sum_{j=1}^s (\mathbf{1}-\mathbf{p}_{i,j}),
\end{equation}
where $s$ is the size of the underpainting patch. Then, performing the $\mathcal{L}^m$ loss when MUI$\ge$0.06. This setting of executing $\mathcal{L}^m$ is set according to the visual effect of the underpainting (demonstrated in Figure \ref{fig-visual}).

The detailed procedure of the proposed UDUP is summarized in Algorithm \ref{alg:algorithm}. Line 4 to line 6 compute loss over mini-batch. 
Considering that Momentum \cite{dong2018boosting} could stabilize the optimization in adversarial attacks. 
Line 10 updates the gradient integrating the momentum to stabilize update directions, which could escape from the possible poor local maxima during the optimization process.

\begin{table*}[t]
	\renewcommand\arraystretch{1.2}
	\centering
	\resizebox{1.7\columnwidth}{!}{
		\begin{tabular}{c|c|c|c|cccl|cc}
			\hline\hline
			\multirow{2}{*}{Method} & \multirow{2}{*}{\begin{tabular}[c]{@{}c@{}}Target\\ Model\end{tabular}} & \multirow{2}{*}{Modification} & \multirow{2}{*}{Universality} & \multicolumn{4}{c|}{Practicality} & \multicolumn{2}{c}{Robustness} \\ \cline{5-10} 
			&  &  &  & Expandable & Arbitrary screen range & Complex background & Plug-and-play & Scaling & JPEG \\ \hline\hline
			\cite{song2018fooling}& STR & Text region & \XSolidBrush & \XSolidBrush & \XSolidBrush & \XSolidBrush &  \XSolidBrush& \XSolidBrush & \XSolidBrush \\
			\cite{chen2020attacking}& STR & Text region & \XSolidBrush & \XSolidBrush &\XSolidBrush  &\XSolidBrush  & \XSolidBrush & \XSolidBrush & \XSolidBrush \\
			\cite{xu2020machines}& STR & Global & \XSolidBrush & \XSolidBrush &\XSolidBrush  &\XSolidBrush  & \XSolidBrush & \XSolidBrush & \XSolidBrush  \\
			\cite{wu2021transferable}& STD & Global & \Checkmark & \XSolidBrush &\XSolidBrush  &\XSolidBrush  & \XSolidBrush & \XSolidBrush & \XSolidBrush \\
			\cite{xiang2022text}& STD & Text region & \XSolidBrush & \XSolidBrush &\XSolidBrush  &\XSolidBrush  & \XSolidBrush & \XSolidBrush & \XSolidBrush \\ \hline
			Our & STD & Underpainting & \Checkmark  & \Checkmark  &\Checkmark   &\Checkmark   &\Checkmark   &\Checkmark   & \Checkmark  \\ \hline\hline
	\end{tabular}}
	\caption{Compared our method with related works from five different aspects. ~"Modification" specifies the range of modification, with ~"Text Region" referring to a local area containing characters. ~"Universality" refers to whether it is suitable for arbitrary characters. ~"Expandable" refers to whether it is suitable for web pages or documents of arbitrary size. ~"Arbitrary screen range" and ~"Complex background" point to whether it is still valid for any screenshot and background with illustrations.}\vspace{-0.5cm}
	\label{tab:Baseline}
\end{table*}

\section{Experimental Results}
\subsection{Experiment Setup}
\noindent \textbf{Models.} Four state-of-the-art STD models, one public OCR tool and two commercial OCR systems are selected to evaluate our defensive strategy, including scene text detectors CRAFT, DBnet, PSENet and PAN++, publicly OCR tool EasyOCR, and off-the-shelf OCRs Aliyun\footnote{\url{https://duguang.aliyun.com}\label{foot-aliyun}} and Baidu\footnote{\url{https://ai.baidu.com/tech/ocr/}}. Except for EasyOCR, these pre-trained models are based on ICDAR2015 \cite{karatzas2015icdar}. All experiments are based on CRAFT to generate universal defensive underpainting patches and conduct black-box evaluations on other models or OCR systems.

\noindent \textbf{Performance Metrics.} The recall rate $\text{R}$ and precision $\text{P}$ are widely used for evaluating the performance of most state-of-the-art STD models, Thus we use them to measure our defense strategy. Considering that the range difference in the results of different models, directly using these metrics will mislead the reader to wrongly estimate the defense performance. Therefore, the ratio values $\text{R}^d/\text{R}^c$ and $\text{P}^d/\text{P}^c$ are used as the evaluation index in the experiment, where $*^d$ and $*^c$ represent the performance of the model before and after adding the defensive underpainting, respectively. We use the mean underpainting intensity (MUI) to measure visual quality according to (\ref{eq-mui}). Although the lower the MUI, the better visual quality is achieved, the visual quality of the image is also affected by the size of the character and patch.

\noindent \textbf{Dataset.} To the best of our knowledge, there is no publicly available dataset that consists of screen-captured text images of various sizes. Therefore, we collected $596$ screenshots of randomly sized English web pages or English documents. Among them, $516$ images are used as the training dataset, and the rest are used as the test set. To cater to the above evaluation criteria, each model has annotated the ground-truth label on a clean dataset. All ground-truth labels are carefully examined to ensure correctness. The dataset and the reproducible data are also released accompanied by the source code.

\noindent \textbf{Hyper-parameters.} In all experiments, we set the step size $\alpha=3/255$, maximum iteration $T=100$, the decay of momentum $\mu=0.1$, perturbation magnitude $\epsilon=30/255$, the hyper-parameters $\beta=6$, the number of mini-batch = $100$. When calculating the dynamic multi-middle-layer loss, we choose the middle four layers: \\ \texttt{basenet\_slice2\_14}, \texttt{basenet\_slice2\_17}, \texttt{upconv1\_conv\_3}, and \texttt{upconv2\_conv\_3}. The settings of balance weight $\lambda$ are $10^{-3}$, $10^{-1}$, $10^{-1}$, $10^{-3}$, $10^{-1}$, $10^{-1}$, and $10^{-2}$ when patch size equal to $10\times10$, $20\times20$, $30\times30$, $50\times50$, $100\times100$, $150\times150$ and $200\times200$. A detailed explanation of the setting of $\lambda$ can be found in the supplementary.

\subsection{Qualitative Comparison with Competing Methods}
This paper focuses on text copyright protection in practical scenarios, considering factors such as universality, visual quality, expandability, and robustness. Specifically, \textit{visual quality} refers to whether the pixels of the characters are modified. To the best of our knowledge, no existing work addresses OCR-resistant text copyright protection in practical scenarios. As shown in Table \ref{tab:Baseline}, we compare our method to the most relevant works from different aspects. The results show that our method is substantially different from previous work. 
The main merit of the proposed UDUP method is that we merely modify the pixels of the underpainting rather than the characters, which makes UDUP easily applied to arbitrary web pages or documents.

\begin{figure*}[th]
	\centerline{\includegraphics[width=0.8\linewidth]{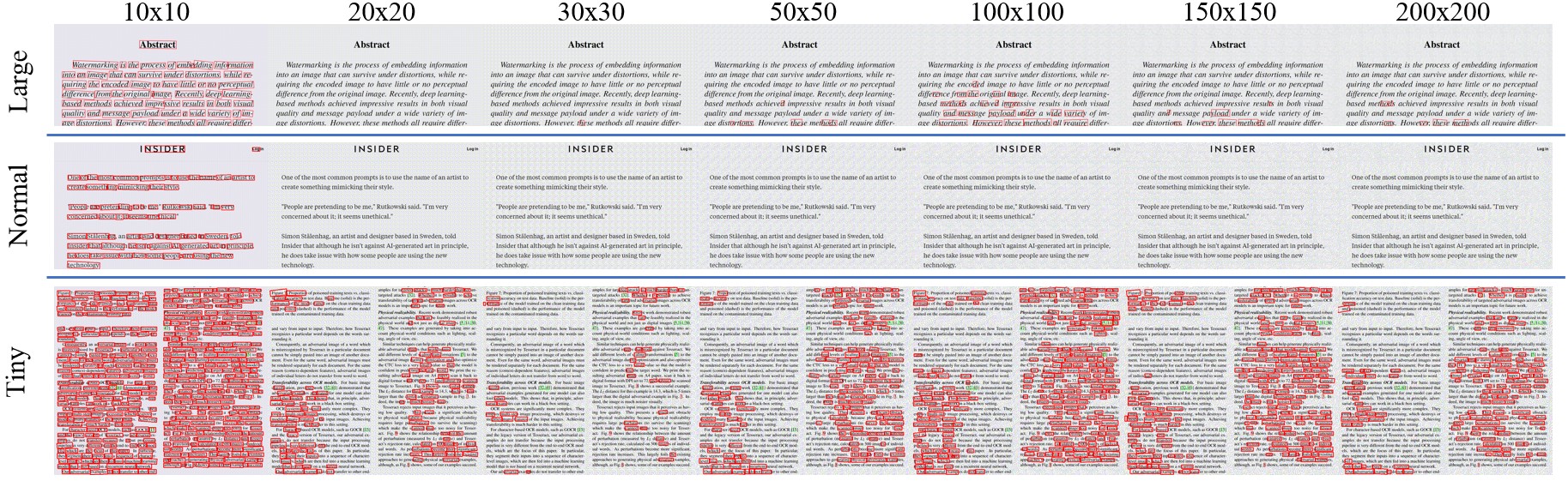}}
	\caption{
		Comparison of the patch size and the defensive effect of UDUP on different types of characters (MUI is fixed as 0.09).}
	\label{fig-ocr-size}
\end{figure*}

\subsection{Visual Quality of UDUP}
To assess the visual quality, this experiment intercepts part of the picture for better observation. As shown in Fig.\ref{fig-visual}, we provide the resultant images after adding different defensive underpainting. One can observe that the interference degree of the underpainting to the vision is directly proportional to the MUI. When MUI=$0.06$, the underpainting is a resembles slight texture, and the entire defensive text image has a clear visual experience. 
When MUI=$0.09$, the influence of underpainting is acceptable and readers can clearly identify each character. When MUI=$0.12$, the background tends to be darker and more chaotic, causing an obvious negative experience to the common human observer, but each character can still be clearly identified.
Therefore we recommend adding defensive underpainting with an MUI of less than $0.12$.
\vspace{-8pt}

\subsection{Universality of The UDUP}
The proposed UDUP distinguishes itself from previous adversarial attacks by only modifying the underpainting. Thus, one may question the effectiveness of UDUP considering that the text image are often consist of different sizes, colors, or types of language for the characters. We in this section conduct a number of experiments to demonstrate the universality of UDUP.

\noindent \textbf{Character-content-agnostic} 
Our test dataset includes screenshots of web pages or documents that feature black characters of different contents and sizes. These screenshots are strictly aligned with real-world scenarios.
Thus, when applying a defensive underpainting to the images within our test dataset, a lower $\text{R}^d$/$\text{R}^c$ score suggests that UDUP provides superior protection against varied character sizes and content. 
Tab.\ref{tab:R} demonstrates UDUP's performance on datasets with different patch sizes and MUIs. 
One can see that, even with an MUI value of 0.06, UDUP remains moderately defensive (except patch size=$10\times10$).  
The weakest defensive strategy (patch size=$100\times100$, MUI=$0.06$) still reduces $\text{R}^d$/$\text{R}^c$ by approximately 34\%. 
Additionally, defense performance is directly proportional to the MUI strength with optimal general defense performance found in patch size=$30 \times 30$, followed by patch sizes of $20 \times 20$, $200\times200$ and $50\times50$. For a setting with patch size=$30/20/50/200$ and MUI=$0.09$, $\text{R}^d$/$\text{R}^c$ equals $0.039/0.059/0.062/0.061$, respectively, making the majority of characters undetectable and disrupting OCR recognition entirely. 
Considering that the defense strategy of patch size $30 \times 30$ possesses the best performance and an MUI value of $0.12$ may subjectively interfere with visual perception, we in the following evaluation focus on the case with the setting of MUI=$0.09$ and patch size= $30\times30$.

\noindent\textbf{Character-size-agnostic.} 
The size of the character is usually optional (\textit{e.g.}, novel website) for some websites. Thus defensive underpainting should be character-size-agnostic. This work classifies text sizes as Large, Normal, and Tiny, based on practical scenarios, and compares the defensive capability of various patch sizes on these three text types. 
The second column of Fig.\ref{fig-ocr-size} exhibits the defensive efficacy of $10\times10$ patches, which is inadequate for providing protection to arbitrary character sizes when the patch size is too small.
The results for the defensive patches ranging from $20$ to $200$ are shown in the second column to the last column. Although these patches effectively protect normal and large-sized characters, they do not provide reasonably good protection for tiny-sized characters.
It is noteworthy that an effective defensive method does not have to facilitate the escape of all characters from detection, but thwarting a portion of them can prevent piracy. 
To this end, one can loosely conclude that UDUP is character-size-agnostic if the patch size$\geq 20$.

\begin{figure}[t]
	\vspace{-8pt}
	\centerline{\includegraphics[width=1.0\linewidth]{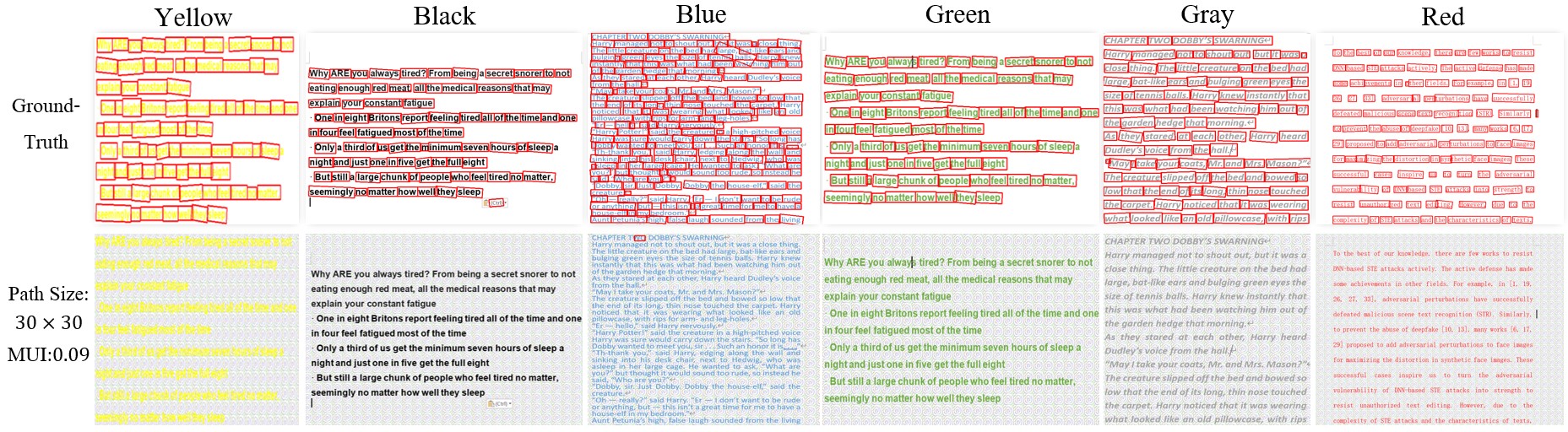}}
	\caption{
		The protection effect of UDUP on different color texts on the CRAFT model.  (patch size=$30\times30$, MUI=0.09)}
	\label{fig-ocr-color}\vspace{-8pt}
\end{figure}

\noindent\textbf{Character-color-agnostic.} 
Typically, in real-world scenarios, text on web pages or documents is not entirely black; some websites even offer an option to change the color of the text.
To evaluate reasonably the independence of the UDUP from the character color, we collected 60 text images and assigned six different colors to them.
The results show that after adding defensive background color, the recall rate of text images with six colors is less than 1\%. 
Fig.\ref{fig-ocr-color} shows the satisfactory protective performance of UDUP against characters in different colors.
Given that the training set is exclusively comprised of black text, we can deduce that UDUP is character-color-agnostic. 

\noindent\textbf{Character-language-agnostic.}
The protective effect of UDUP on Chinese, English, Arabic, and Japanese languages is demonstrated in Fig.\ref{fig-real}. Despite the fact that the training dataset for the underpainting patch in the experiment is exclusively in English, the results can be extrapolated to other languages.
\vspace{-8pt}

\begin{table}[t]
	\renewcommand\arraystretch{1.1}
	\centering
	\resizebox{0.8\columnwidth}{!}{
		\begin{tabular}{cccccccc}
			\hline 			\hline
			\multirow{2}{*}{\begin{tabular}[c]{@{}c@{}}Patch\\ Size\end{tabular}} & \multicolumn{7}{c}{MUI} \\ \cline{2-8} 
			& $0.06$ & $0.07$ & $0.08$ & $0.09$ & $0.10$ & $0.11$ & $0.12$ \\ \hline \hline
			$10$ & $0.864$ & $0.815$ & $0.741$ & $0.571$ & $0.260$ & $0.162$ & $0.083$ \\
			$20$ & $0.249$ & $0.141$ & $0.094$ & $0.059$ & $0.026$ & $0.013$ & $0.005$ \\
			$30$ & $\textbf{0.209}$ & $\textbf{0.121}$ & $\textbf{0.080}$ & $\textbf{0.039}$ & $\textbf{0.014}$ & $\textbf{0.002}$ & $\textbf{0.001}$ \\
			$50$ & $0.430$ & $0.219$ & $0.115$ & $0.062$ & $0.026$ & $0.007$ & $0.002$ \\
			$100$ & $0.669$ & $0.260$ & $0.138$ & $0.052$ & $0.038$ & $0.019$ & $0.006$ \\
			$150$ & $0.533$ & $0.263$ & $0.117$ & $0.066$ & $0.026$ & $0.008$ & $0.003$ \\
			$200$ & $0.341$ & $0.180$ & $0.107$ & $0.061$ & $0.027$ & $0.012$ & $0.003$ \\ \hline \hline
		\end{tabular}
	}
	\caption{The value of $\text{R}^d/\text{R}^c$ of the test dataset on CRAFT after adding UDUP with different patch sizes and MUIs. Smaller value $\text{R}^d/\text{R}^c$ indicates better performance.}\vspace{-8pt}
	\label{tab:R}
\end{table}

\subsection{Application to Real-world Scenarios}
UDUP has been shown in the previous section to have sufficient generality under ideal conditions (\emph{i.e.}, pure color background and properly limited screenshot area). However, real-world scenarios are complex and may encounter the following issues. First, the extent of the screenshot is unpredictable. It means that UDUP must be effective in arbitrary screenshot range. Second, documents and web pages usually contain a wide variety of illustrations, so the presence of these images in the screenshot area could also compromise the defensive effect of UDUP. Therefore, this section will examine the effectiveness of UDUP in real-world scenarios.

\begin{figure}[]
	\vspace{-8pt}
	\captionsetup[subfigure]{labelformat=empty}
	\centering
	\subfloat[Chinese]{\label{figure_caaa}\includegraphics[width=0.12\textwidth]{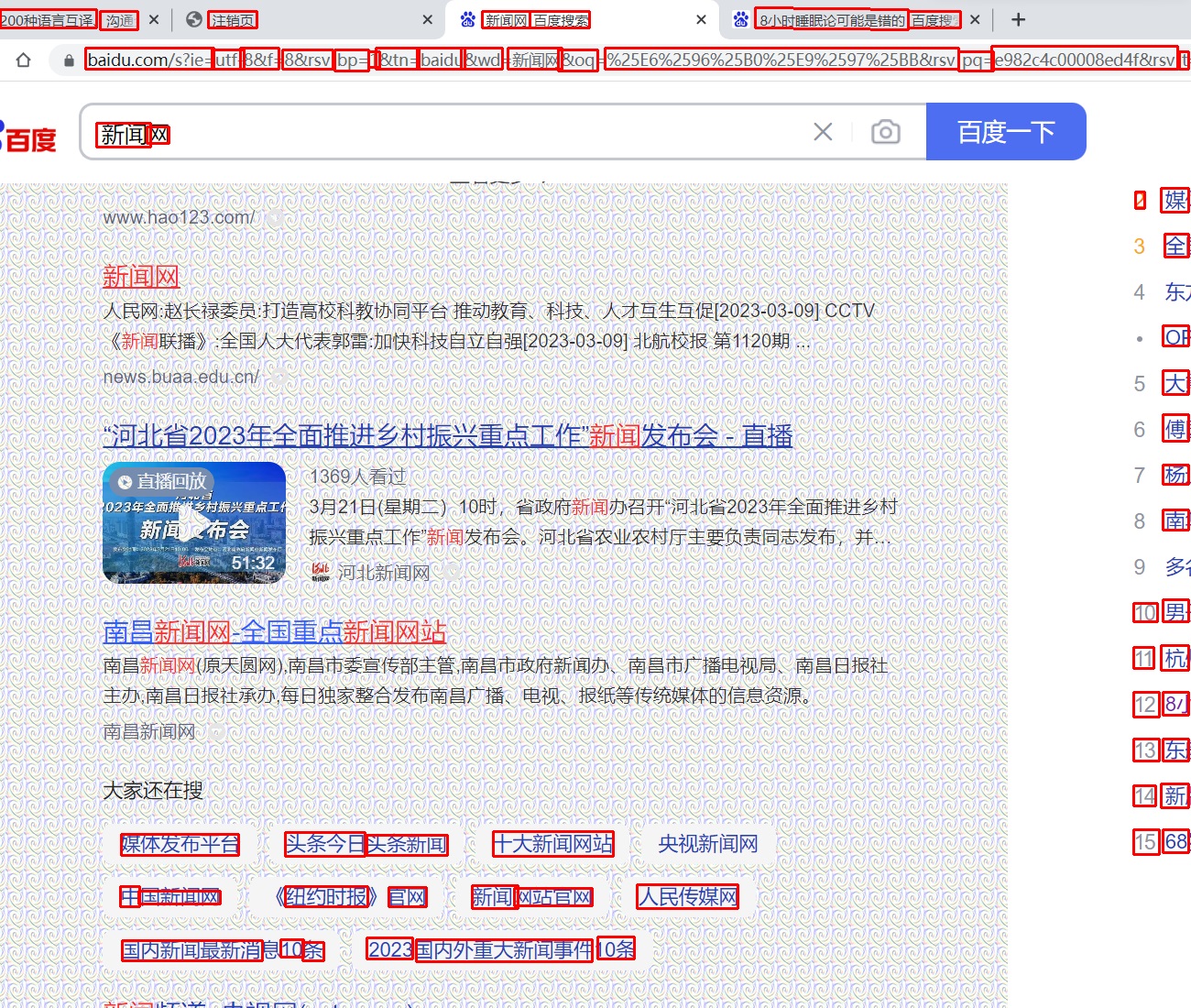}}
	\subfloat[Arabic]{\label{figure_cbbb}\includegraphics[width=0.12\textwidth]{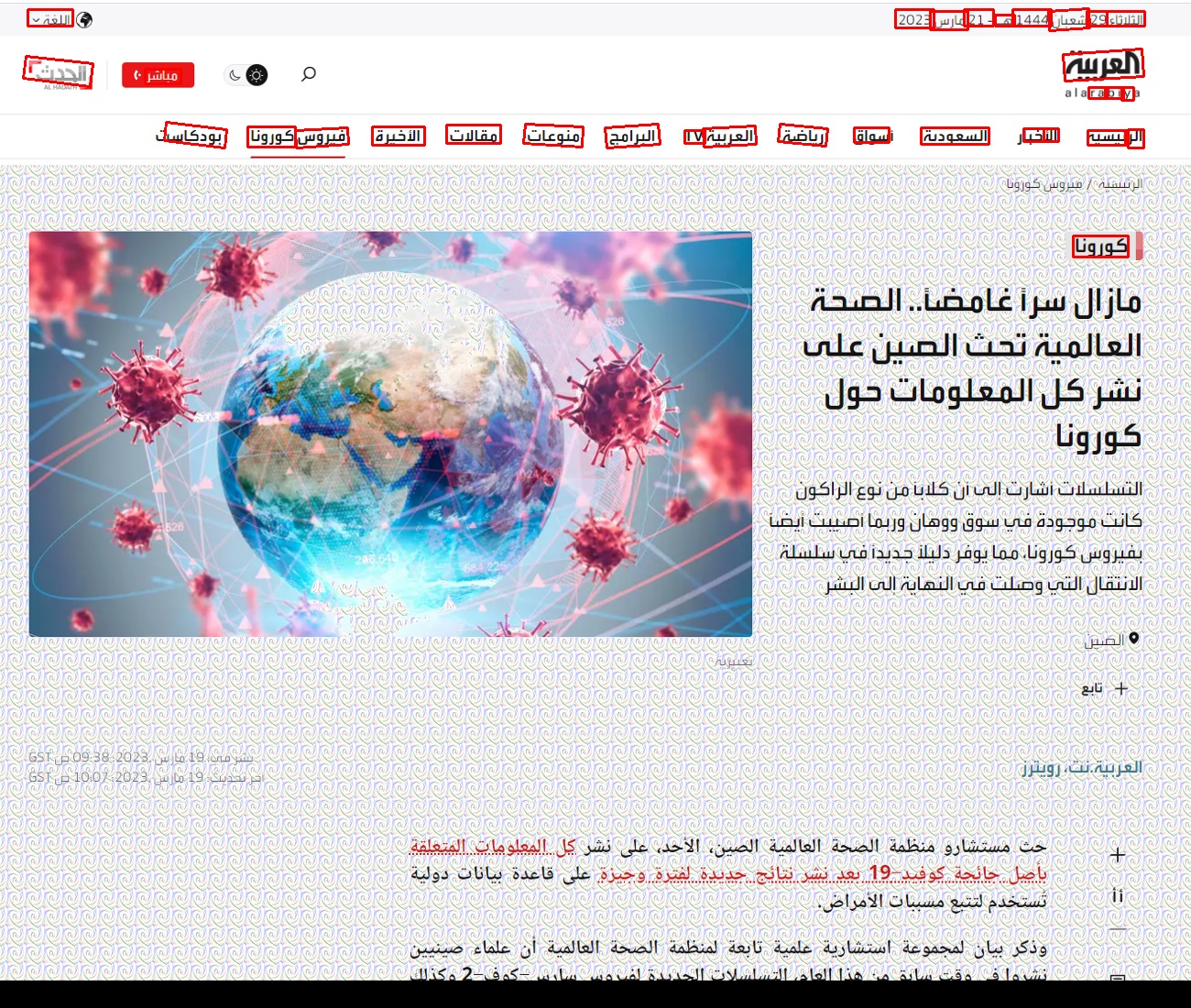}}
	\subfloat[English]{\label{figure_cccc}\includegraphics[width=0.12\textwidth]{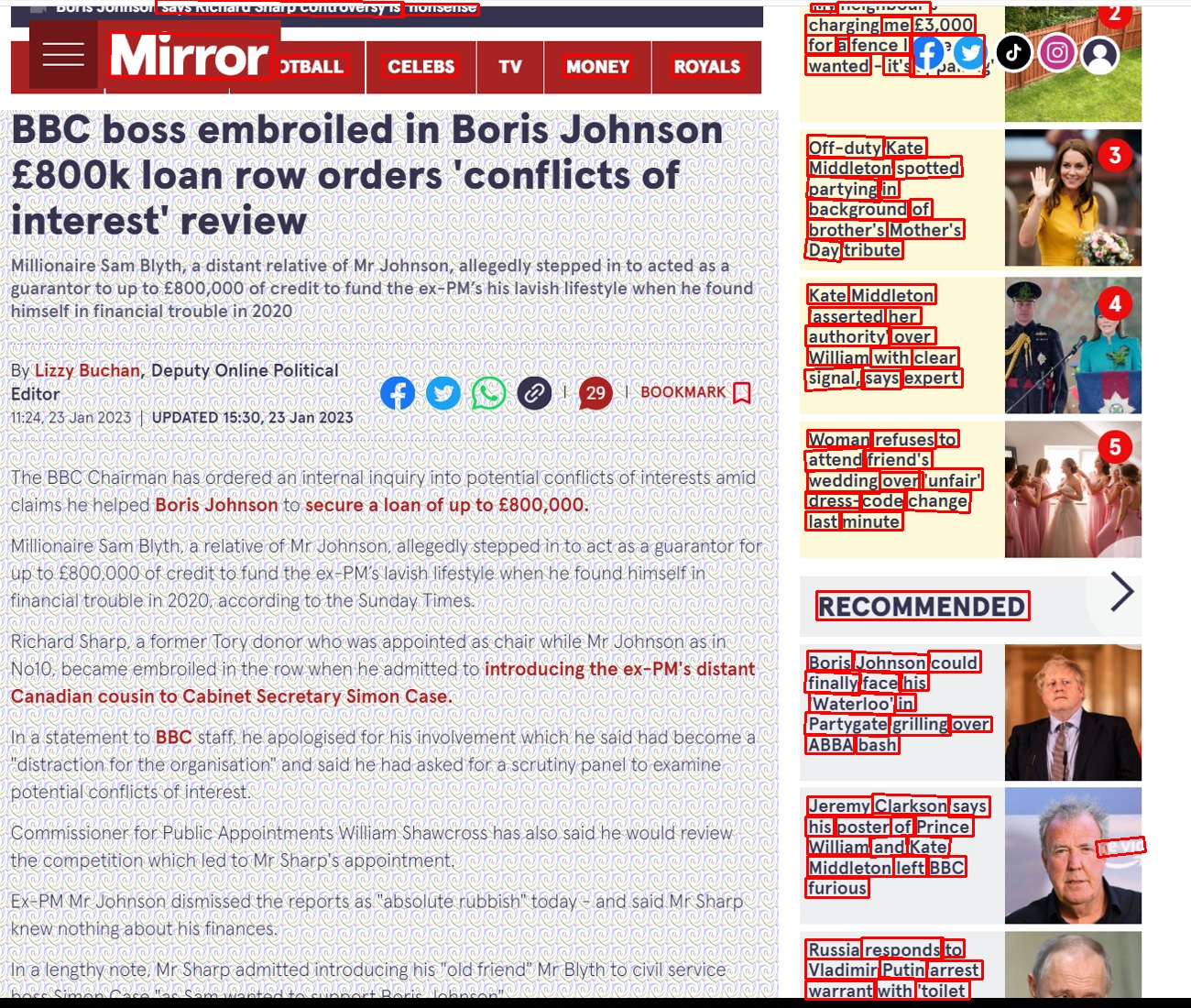}}
	\subfloat[Japanese]{\label{figure_cddd}\includegraphics[width=0.12\textwidth]{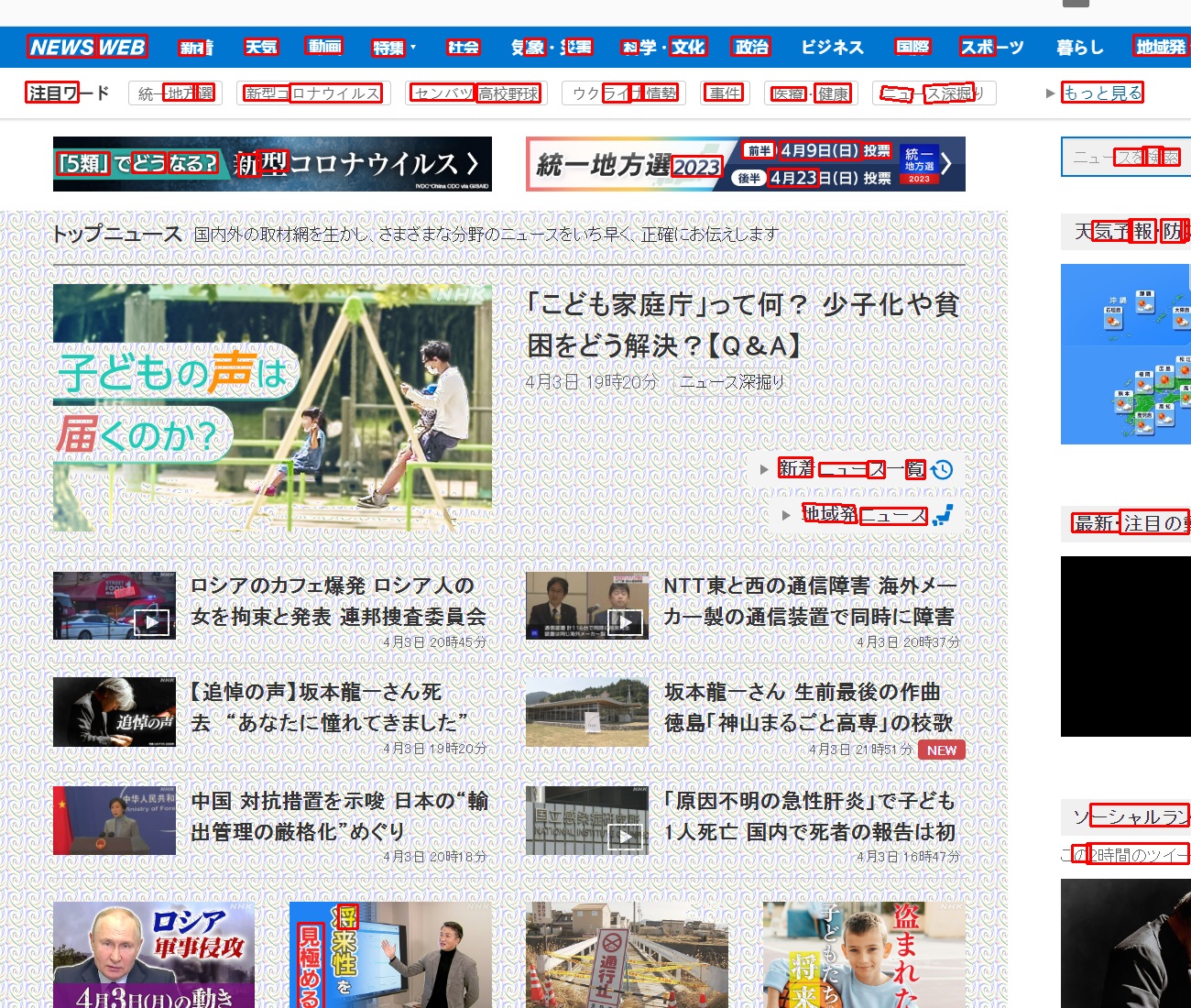}}
	\caption{Four real-world cases featuring a variety of contents, colors, fonts, sizes, languages, arbitrary image-based backgrounds, and arbitrary screenshot sizes. Red boxes indicate the detection of CRAFT. The majority of characters over UDUP underpainting cannot be detected. Zoom in for better visualization. (Patch size=$30\times30$, MUI=0.09)}
	\label{fig-real}
	\vspace{-8pt}
\end{figure}

Fig.\ref{fig-real} shows four screenshots of news websites with a large number of illustrations as examples and replaced the background of the central area of the webpage with a defensive underpainting to simulate arbitrary screenshot areas. These screenshots feature a wide range of characters, including different colors, sizes, fonts, and languages. The experimental results demonstrate that UDUP performs remarkably well even in real-world scenarios. The majority of characters beyond the underpainting area were precisely located, while those within that area were undetected. Notably, our training set comprised exclusively of text images featuring a solid color background and text filled with the underpainting patch. Our study supports the conclusion that UDUP can be deployed in real-world scenarios where complex graphics are present, and it can operate reliably within screenshots of arbitrary ranges.

\begin{table}[]
	\centering
	\renewcommand\arraystretch{1.2}
	\resizebox{0.9\columnwidth}{!}{
		\small
		\begin{tabular}{c|cccccc|l}
			\hline\hline
			\multirow{2}{*}{\begin{tabular}[c]{@{}c@{}}Patch\\ Size\end{tabular}} & \multicolumn{6}{c|}{JPEG Quality Factor($Q$)} & \multirow{2}{*}{None} \\ \cline{2-7}
			& $50$ & $60$ & $70$ & $80$ & $90$ & $100$ &  \\ \hline\hline
			$30\times30$ & $0.209$ & $0.171$ & $0.140$ & $0.112$ & $0.096$ & $0.085$ & $0.039$ \\ \hline\hline
		\end{tabular}
	}
	\caption{$\text{R}^d/\text{R}^c \downarrow$ at different JEPG qualities factors on CRAFT.}\vspace{-0.35cm}
	\label{tab:jpeg}	
\end{table}

\section{ROBUSTNESS OF UDUP}
The vulnerability of adversarial examples makes them susceptible to attacks by operations such as scaling and compression, posing a challenge to their effectiveness. UDUP, operating on similar adversarial attack principles, is likewise susceptible to these risks. We in this section evaluate the robustness of UDUP against typical image manipulations such as scaling, and image compression (\emph{e.g.}, JPEG).

\noindent\textbf{Robustness to image scaling.}
While scaling web pages or documents is a common operation, it can undermine the defensive effectiveness of UDUP. Thanks to the explicit incorporation of the scaling module, the proposed UDUP is resilient to scaling, with the experiment conducted over a range of scaling ratios from 60\% to 200\%.
Smaller scaling factors were not considered due to their negative impact on OCR precision. More details on the setting of this experiment are given in the supplementary. We constructed a dataset comprising of 20 text image screenshots of varying sizes from web pages and documents.
Fig.\ref{fig-ocr-rescale} presents the defensive performance of UDUP after scaling. 
One can see that the zoom-out operation slightly hampers the defensive performance of UDUP, especially when MUI=$0.09$ and scaling factor=$0.6$. 
The zoom-in operation still retains acceptable defensive performance.
However, even in the worst-case (scale factor = $0.6$, MUI = $0.09/0.12$), our suggested defensive underpainting can still reduce $\text{R}^d$/$\text{R}^c$ by approximately 71\%/94\%, validating that UDUP is robustness to scaling.

\noindent\textbf{Robustness to JPEG.}
Tab.\ref{tab:jpeg} shows the influence of JPEG quality factors on UDUP (MUI=0.09), with the last column indicating no JPEG compression. With the decrement in quality factor, the defensive capability of UDUP drops. However, for low-quality JPEG, \emph{e.g.}, quality factor (Q) is 50, the decrement of UDUP defence is also acceptable and the corresponding $\text{R}^d$/$\text{R}^c$ is 0.209, the robustness to JPEG may due to that the UDUP underpainting is mainly composed of low-frequency perpetuation.

\begin{table}[]
	\centering
	\renewcommand\arraystretch{1.2}
	\resizebox{0.9\columnwidth}{!}{
		\small
		\begin{tabular}{cccccc}
			\hline \hline
			\multirow{2}{*}{MUI} & \multicolumn{5}{c}{Model} \\ \cline{2-6} 
			& CRAFT*\cite{Baek2019CVPR} & DBNet & EasyOCR & PAN++\cite{wang2021pan++} & PSENet\cite{wang2019shape} \\ \hline \hline
			\cellcolor{lightgray}0 & \multicolumn{5}{c}{\cellcolor{lightgray}1.00/1.00} \\ \hline
			0.09 & 0.03/0.75 & 0.83/0.23 & 0.37/0.37 & 0.38/0.78 & 0.55/0.79 \\
			0.12 & 0.00/0.60 & 0.68/0.26 & 0.17/0.26 & 0.08/0.73 & 0.30/0.66 \\ \hline \hline
		\end{tabular}
	}
	\caption{$\text{R}^d/\text{R}^c\downarrow$ and $\text{P}^d/\text{P}^c\downarrow$ of patch sizes=$30\times30$ for scene text detection models. The superscript * of CRAFT indicates the model is tested under the white-box setting and the others are in black-box settings.}
	\label{tab:transferability}
	\vspace{-8pt}
\end{table}

\begin{figure}[t]
	\vspace{-20pt}
	\captionsetup[subfigure]{}
	\centering
	\subfloat[DBNet]{\includegraphics[width=0.16\textwidth]{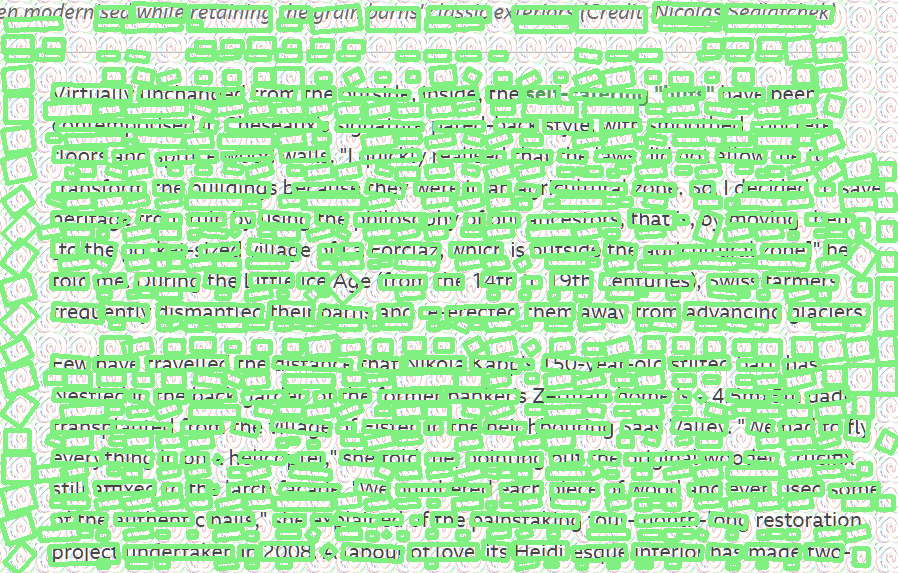}\label{fig-tab-dbnet}}
	\subfloat[EasyOCR]{\includegraphics[width=0.16\textwidth]{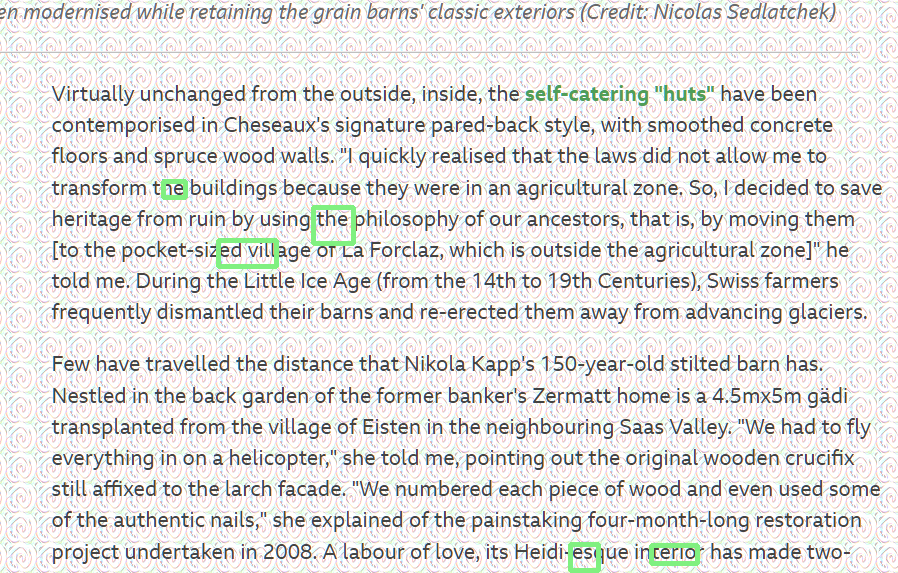}}
	\subfloat[PAN++]{\includegraphics[width=0.16\textwidth]{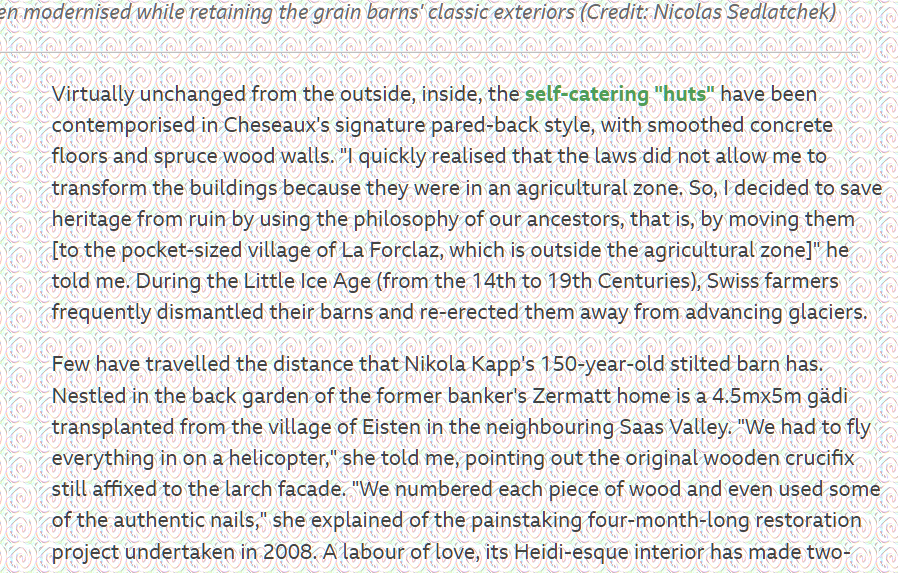}}\\
	\vspace{-8pt}
	\subfloat[PSENet]{\includegraphics[width=0.16\textwidth]{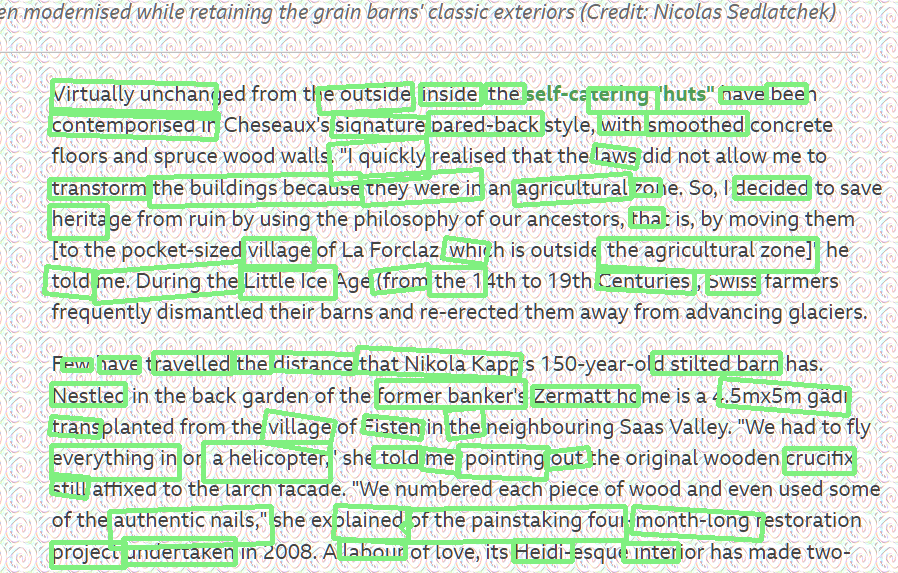}}
	\subfloat[Aliyun]{\includegraphics[width=0.16\textwidth]{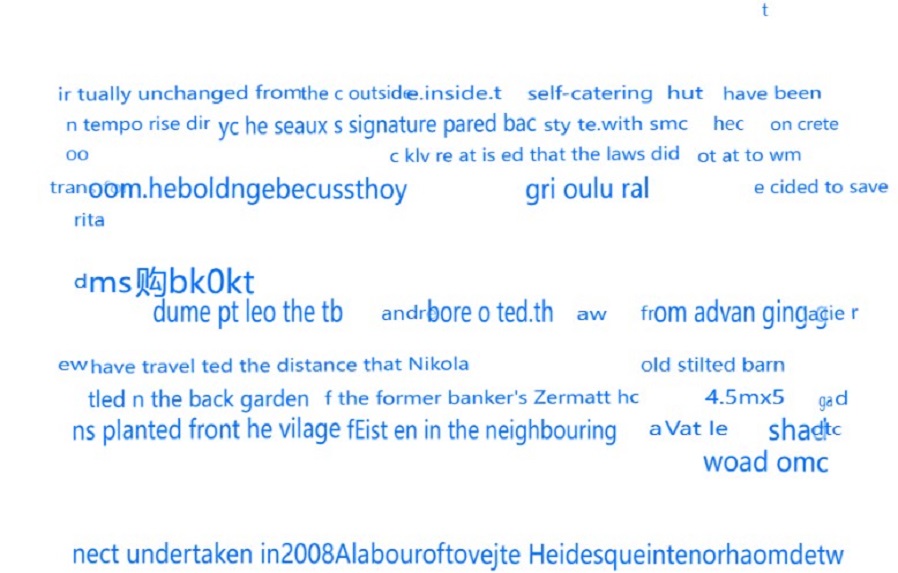}}
	\subfloat[Baidu]{\includegraphics[width=0.16\textwidth]{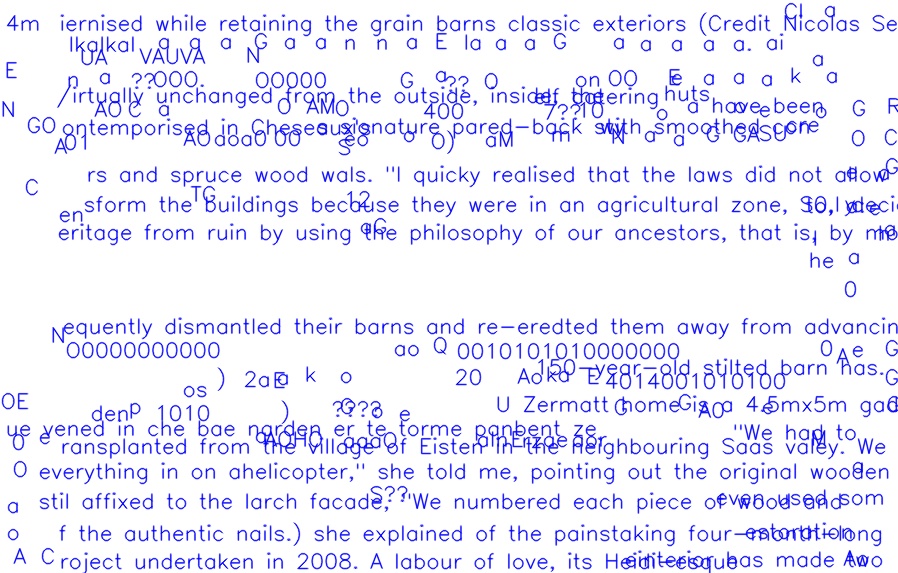}\label{fig-tab-baidu}}
	\caption{The recognition results of adding UDUP on black-box models.}
	\label{fig-ocr-transferability}\vspace{-8pt}
	
\end{figure}

\section{Transferability of UDUP}
The transferability of adversarial examples refers to the property that adversarial examples crafted for one model can often be used to cause misclassification of other models, even if the models were trained on different data and constructed with different architectures. In real-world situations, it is difficult to anticipate which OCR system will be used by pirates, hence the significance of examining the transferability of UDUP across various OCR systems. In this section, we evaluate the transferability of UDUP, including four scene text detectors and two off-the-shelf commercial OCR systems.

\noindent\textbf{Transferability on black-box STDs.}
As shown in Tab.\ref{tab:transferability}, under the settings of MUI=$0.09$ and $0.12$, UDUP significantly reduce the $\text{R}^d$/$\text{R}^c$ on other black-box models (except DBNet). Although the protective effect is not perfect for white-box models, UDUP's transferability has been proven. 
For DBnet, although the recall rate is not significantly reduced, the precision rate is reduced to about 25\%. Note that reducing precision also protects text copyright. As shown in Fig.\ref{fig-tab-dbnet}, a large number of areas are mislocated by DBNet and these areas could be recognized as text (\textit{e.g.}, Fig.\ref{fig-tab-baidu}). Text copyrights can be protected when the result is a large amount of confusing or irrelevant content. This phenomenon indicates that UDUP permits the underpainting to generate character features, making the model unable to distinguish characters from the underpainting correctly. 

\noindent\textbf{Transferability on commercial OCR systems.}
As illustrated in Fig.\ref{fig-ocr-transferability}, we evaluated the effectiveness of UDUP against black-box commercial OCR systems.
One can find that while both Aliyun and Baidu OCR locates most of the characters correctly, there are still some characters that are not located correctly.
Furthermore, it's observed that both Aliyun Cloud OCR and Baidu OCR have numerous spelling errors in their detection results, and Baidu OCR recognizes a large number of blank areas as text.
These results suggest that UDUP can effectively protect characters from detection by commercial OCR systems.
In addition, we also notice a by-product function of the proposed UDUP: as an effective measure to evaluate the robustness of the model fairly. Its fairness roots in the fact that UDUP only modifies the underpainting of text images, not the text itself, which is strictly in line with the complex text background of the real-world scenarios.

\begin{table}[t]
	\renewcommand\arraystretch{1.1}
	\resizebox{0.9\columnwidth}{!}{
		\small
		\begin{tabular}{c|ccccc}
			\hline \hline
			\multirow{2}{*}{$\mathcal{L}^p/\mathcal{L}^m/R_1$} & \multicolumn{5}{c}{$\text{R}^d$/$\text{R}^c\downarrow$ and $\text{P}^d$/$\text{P}^c\downarrow$ on different models} \\ \cline{2-6} 
			& CRAFT & DBNeT & EasyOCR & PAN++ & PSENet \\ \hline \hline
			\cellcolor{lightgray}{clean} & \multicolumn{5}{c}{\cellcolor{lightgray}1.00/1.00} \\ \hline
			\Checkmark/\XSolidBrush/\XSolidBrush & 0.15/0.75 & 0.90/0.39 & 0.52/0.51 & 0.22/0.74 & 0.65/0.81 \\ \cline{1-1}
			\Checkmark/\XSolidBrush/\Checkmark & 0.08/0.76 & 0.88/0.35 & 0.46/0.47 & 0.16/0.72 & 0.59/0.78 \\ \cline{1-1}
			\Checkmark/\Checkmark/\XSolidBrush & 0.10/0.73 & 0.86/0.35 & 0.48/0.48 & 0.16/0.71 & 0.61/0.79 \\ \cline{1-1}
			\Checkmark/\Checkmark/\Checkmark & 0.03/0.75 & 0.83/0.23 & 0.37/0.37 & 0.08/0.73 & 0.55/0.79 \\  \cline{1-1} \hline \hline
		\end{tabular}
	}
	\caption{Ablation studies on multi-middle-loss $\mathcal{L}^m$ and random scaling module $R_1$. (MUI=$0.09$ and patch size= $30\times30$)}
	\label{tab:ablation}
	\vspace{-10pt}
\end{table}

\begin{figure}[t]
	\vspace{-8pt}
	\centerline{\includegraphics[width=0.8\linewidth]{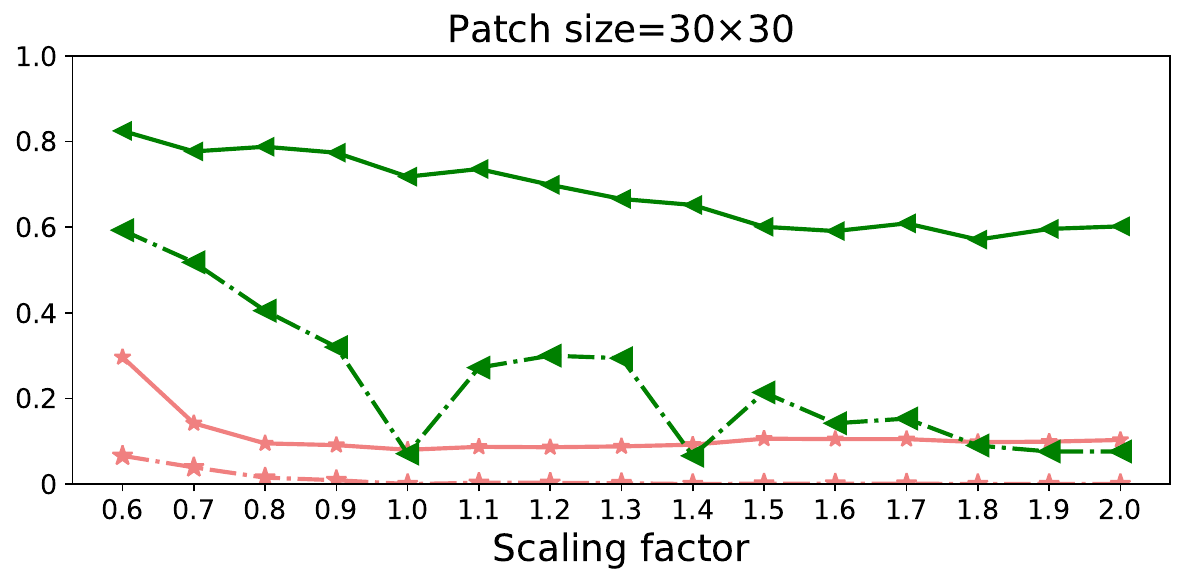}}
	\caption{
		$\text{R}^d$/$\text{R}^c\downarrow$ and $\text{P}^d$/$\text{P}^c\downarrow$ v.s. scaling factor under different settings of UDUP on CRAFT. The green line and the red line represent $\text{P}^d$/$\text{P}^c$ and $\text{R}^d$/$\text{R}^c$, respectively. Dotted lines and solid lines represent the setting of MUI=$0.12$ and MUI=$0.09$.}
	\label{fig-ocr-rescale}
	\vspace{-8pt}
\end{figure}

\section{Ablation Study}
We analyze the effectiveness of loss $\mathcal{L}^m$ and random scaling module $R_1$. As shown in Tab.\ref{tab:ablation}, ablation experiments were conducted under the setting of patch size=$30\time30$ and MUI=$0.09$. One can find that incorporating both $\mathcal{L}^m$ and $R_1$ further reduces the recall $\text{R}^d$/$\text{R}^c$. Similarly, the precision $\text{P}^d$/$\text{P}^c$ of DBNet and EasyOCR decline as well. This demonstrates that the proposed $\mathcal{L}^m$ and $R_1$ can effectively enhance the defensive capability of UDUP. Due to space limitations, we provide the $\text{R}^d$/$\text{R}^c$ results for various balancing weights $\lambda$ in the supplementary.

\section{Conclusion}
This work aims to devise a universal defensive underpainting patch (UDUP) that can safeguard characters from unlawful detection by optical character recognition (OCR) systems in any screenshot range and complex background. The experimental results under both white-box and black-box settings demonstrate that the proposed UDUP is capable of adapting to the text of different sizes, colors, fonts, and languages, and simultaneously maintaining visual quality and transferability to black-box models. Moreover, the proposed UDUP is robust against typical image processes including scaling, and image compression. UDUP could find many practical applications such as safeguarding text copyright, text captcha, and evaluating OCR robustness.

\section*{Acknowledgments}

This work was supported by the National Natural Science Foundation of China (Grant No. 61901237, 62171244, 62072343, 62202009), Zhejiang Provincial Natural Science Foundation of China (Grant No. LY23F020011), Alibaba Innovative Research.

\bibliographystyle{ACM-Reference-Format}
\balance
\bibliography{ref.bib}

\newpage
\appendix
\section{Appendix}
In this supplementary materiel, we provide more results on 1) testing on a variety of rescaling ratios, 2) the impact on the choice on the hyperparameter $\lambda$.

\subsection{Rescaling Ratios}
In this section, we supplement the basis for setting the rescaling range in the robustness experiments. In the manuscript, the rescaling factor ranges from 60\% to 200\%. This is because the too-small text will affect the accuracy of OCR. The specific proof experiment is as follows. First, we edit some text with Microsoft Word software in font size 10.5. Then the document is rescaled at different rescaling ratios, and the screenshots are captured. Finally, these screenshots are sent to online commercial OCRs, including Tencent OCR and Aliyun OCR\footref{foot-aliyun}. Fig.1 illustrates that the result obtained from Aliyun OCR contains errors when the rescaling ratio is set to 60\%. Additionally, the results obtained at rescaling ratios of 50\% and 40\% exhibit significant recognition errors. Therefore, the lower limit of the rescaling ratio is 60\%.

\begin{figure}[h]
	\centerline{\includegraphics[width=1.0\linewidth]{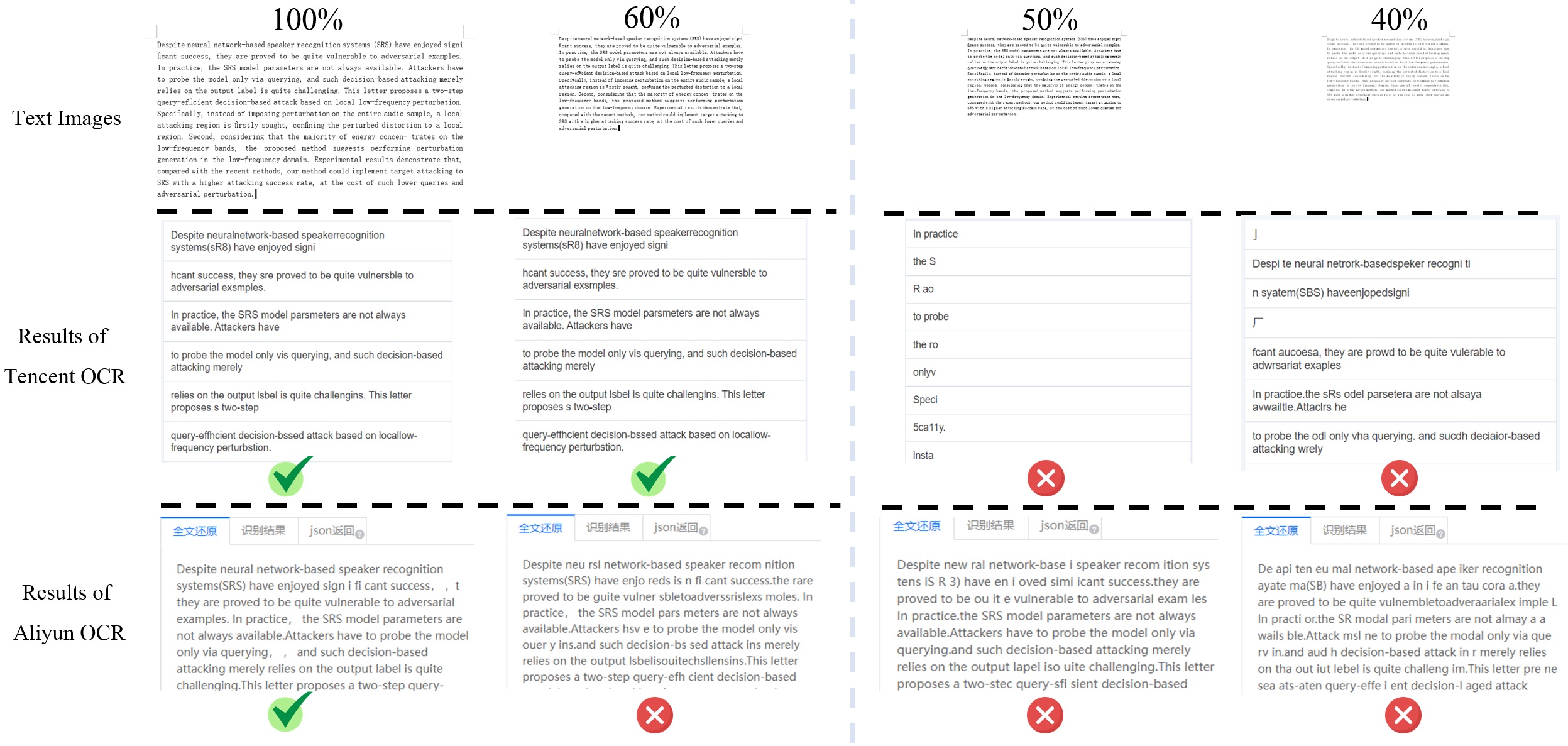}}
	\caption{A comparison of the results of commercial OCR under various scaling ratios, with \Checkmark representing the accurate recognition and \XSolidBrush indicating an recognition with numerous errors. To facilitate easy comparison, only the head results of the commercial OCR results are posted.}
	\vspace{-5pt}
	\label{fig-ocr-rescale-ratio}
\end{figure}

\subsection{The Setting of Hypermeter $\lambda$}
This section supplement the ablation experiment about balance weight $\lambda$ on the white-box CRAFT. As shown in Fig.\ref{fig-ocr-rescale-ratio}, the preset loss function $\mathcal{L}^*$ in the full paper includes prediction loss $\mathcal{L}^p$ and multi-middle-layer loss $\mathcal{L}^m$. The hypermeter $\lambda$ denotes the balance weights.

\begin{equation}\label{equ:2}
	\mathcal{L}^*_t=\underset{(\mathbf{x},\mathbf{M}) \sim \mathcal{N}}{\mathbb{E}}[ \mathcal{L}^p(\mathbf{x},\mathbf{p}_t)+\lambda \mathcal{L}^m(\mathbf{x},\mathbf{p}_t)],
\end{equation}

Then we evaluate the recall ratio $\text{R}^d/\text{R}^c$ under the different settings of $\lambda$. 
As shown in the Table \ref{tab:ablation} , we emphasize in bold the best results when using different $\lambda$ under different patch sizes. It can be found that the defensive effect of UDUP after adding appropriate hyper-parameters is always better than that when $\lambda$ is equal to $0$. Taking patch size=$20\times20$ as an example, when $\lambda$ is equal to $10^{-1}$, the defense effect is increased by $0.175$ compared with $\lambda=0$. 
Hence, it can be concluded that the loss function $\mathcal{L}^m$ is successful in improving the protection capabilities of UDUP. 
At the same time, we also determined the settings of $\lambda$ under different patch sizes. The settings of balance weight $\lambda$ are $10^{-3}$, $10^{-1}$, $10^{-1}$, $10^{-3}$, $10^{-1}$, $10^{-1}$, and $10^{-2}$ when patch size equal to $10\times10$, $20\times20$, $30\times30$, $50\times50$, $100\times100$, $150\times150$ and $200\times200$.

\begin{table}[h]
	\renewcommand\arraystretch{1.2}
	\resizebox{0.8\columnwidth}{!}{
		\small
		\begin{tabular}{ccccccc}
			\hline 		\hline
			\multirow{2}{*}{\begin{tabular}[c]{@{}c@{}}Patch\\  Size\end{tabular}} & \multicolumn{6}{c}{$\lambda$} \\ \cline{2-7} 
			& \cellcolor{lightgray}$0$ & $10^{-3}$ & $10^{-2}$ & $10^{-1}$ & $10^{0}$ & $10^1$ \\ \hline 		\hline
			$10$ & \cellcolor{lightgray} $0.795$ & $\textbf{0.571}$ & $0.840$ & $0.749$ & $0.720$ & $0.735$  \\
			$20$ & \cellcolor{lightgray}$0.234$ & $0.634$ & $0.007$ & $\textbf{0.059}$ & $0.446$ & $0.375$  \\
			$30$ & \cellcolor{lightgray}$0.083$ & $0.077$ & $0.081$ & $\textbf{0.039}$ & $0.108$ & $0.121$  \\
			$50$ & \cellcolor{lightgray}$0.078$ & $\textbf{0.062}$ & $0.075$ & $0.095$ & $0.094$ & $0.102$  \\
			$100$ & \cellcolor{lightgray}$0.065$ & $0.058$ & $0.063$ & $\textbf{0.052}$ & $0.067$ & $0.072$  \\
			$150$ & \cellcolor{lightgray}$0.081$ & $0.066$ & $0.174$ & $\textbf{0.066}$ & $0.069$ & $0.075$ \\
			$200$ & \cellcolor{lightgray}$0.074$ & $0.062$ & $\textbf{0.061}$ & $0.072$ & $0.067$ & $0.071$ \\ \hline		\hline
		\end{tabular}
	}
	\caption{The ablation study on loss functions. We show the $\text{R}^d$/$\text{R}^c$ for UDUP with different patch sizes and $\lambda$ on CRAFT. Lower $\text{R}^d$/$\text{R}^c$ values indicate a stronger defensive effect. The gray column represents the results without loss function $\mathcal{L}^m$. (MUI fixed as $0.09$)}
	\label{tab:ablation}
\end{table}

\end{document}